\pdfoutput=1

\documentclass[11pt]{article}

\usepackage[final]{acl}

\usepackage{times}
\usepackage{latexsym}

\usepackage[T1]{fontenc}

\usepackage[utf8]{inputenc}

\usepackage{microtype}

\usepackage{inconsolata}

\usepackage{bibentry}

\usepackage[utf8]{inputenc}
\usepackage{subfigure}
\usepackage{booktabs}
\usepackage{amsmath}
\usepackage{amsfonts}
\usepackage{multirow}
\usepackage{color}
\usepackage{enumitem}
\usepackage{microtype}
\usepackage{tabularx}

\usepackage{chngpage}
\usepackage{array}
\usepackage{microtype}
\usepackage{xcolor}
\usepackage{fontenc}
\usepackage{arydshln}
\usepackage{array}
\usepackage{booktabs}

\usepackage{newfloat}
\usepackage{listings}
\usepackage{makecell}
\usepackage{aliascnt}

\usepackage{algorithm}
\usepackage{algorithmic}
\usepackage{hyperref}
\usepackage{graphicx}
\usepackage{amssymb}

\usepackage[utf8]{inputenc}
\usepackage{cleveref}
\crefname{section}{§}{§§}
\Crefname{section}{§}{§§}

\title{Dual Instruction Tuning with Large Language Models for \\ Mathematical Reasoning}

\author{Yongwei Zhou, Tiejun Zhao\\
    Harbin Institute of Technology, Harbin, China\\
    ywzhouphd2018@gmail.com \;\;  tjzhao@hit.edu.cn
    }

\begin{document}
\maketitle
\begin{abstract}
Recent advancements highlight the success of instruction tuning with large language models (LLMs) utilizing Chain-of-Thought (CoT) data for mathematical reasoning tasks. 
Despite the fine-tuned LLMs, challenges persist, such as incorrect, missing, and redundant steps in CoT generation leading to inaccuracies in answer predictions. 
To alleviate this problem, we propose a dual instruction tuning strategy to meticulously model mathematical reasoning from both forward and reverse directions.
This involves introducing the Intermediate Reasoning State Prediction task (forward reasoning) and the Instruction Reconstruction task (reverse reasoning) to enhance the LLMs' understanding and execution of instructions. 
Training instances for these tasks are constructed based on existing mathematical instruction tuning datasets. 
Subsequently, LLMs undergo multi-task fine-tuning using both existing mathematical instructions and the newly created data.
Comprehensive experiments validate the effectiveness and domain generalization of the dual instruction tuning strategy across various mathematical reasoning tasks.
\end{abstract}

\section{Introduction}
\begin{figure*}[t]
	\centering
	\includegraphics[width=6.2in]{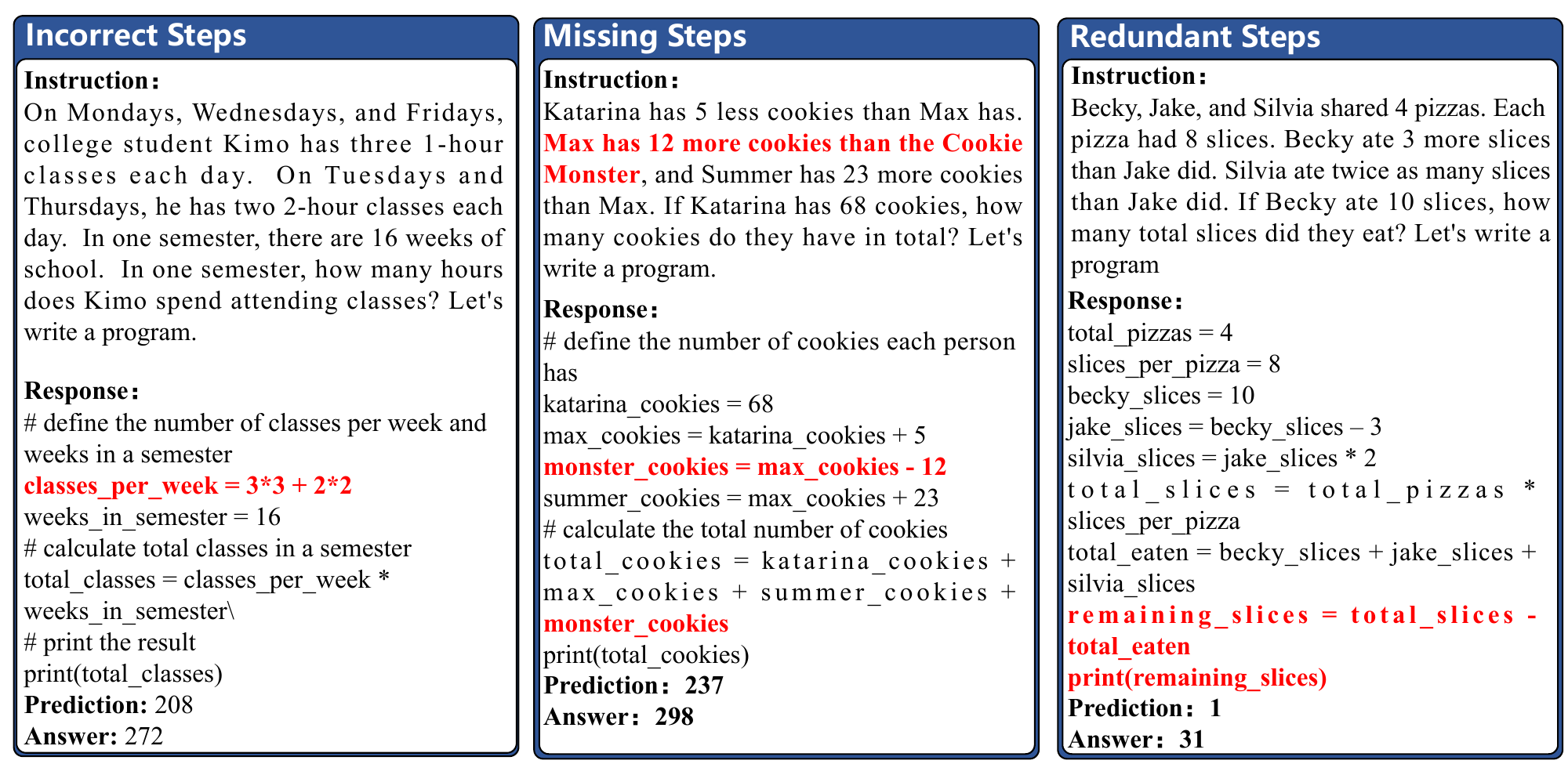}
	\caption{The errors, omissions, and redundancies in the generation process of reasoning steps}
    \label{fig:errors}
    \vspace{-0.2cm}
\end{figure*}

With the advance of large language models (LLMs)~\cite{DBLP:journals/corr/abs-2302-13971,DBLP:journals/corr/abs-2308-12950,DBLP:conf/acl/DuQLDQY022,DBLP:journals/corr/abs-2303-08774,DBLP:journals/corr/abs-2212-08073,DBLP:journals/corr/abs-2302-13971} and Chain-of-Thought (CoT) techniques~\cite{DBLP:journals/corr/abs-2201-11903,DBLP:journals/corr/abs-2302-12246,DBLP:journals/corr/abs-2211-12588}, the mathematical reasoning ability of models have significantly been improved, particularly in zero-shot and few-shot scenarios.
As illustrated in three examples in Figure \ref{fig:errors}, due to the inadequacy of understanding and execution of instruction, LLMs still exhibit errors, omissions, and redundancies in generating thoughts, leading to incorrect final results.

To alleviate this issue, we propose a dual instruction tuning strategy to meticulously model mathematical reasoning from both forward and reverse directions.
Specifically, this strategy involve two auxiliary tasks, including Intermediate Reasoning State Prediction (IRSP) and the Instruction Reconstruction (IR) task.
The former is prompted to predict the masked reasoning steps (forward reasoning) based on the provided context and partially revealed reasoning steps, while the later randomly masks clauses or questions containing numerical values in instructions and prompts to reconstruct them (reverse reasoning).
Training instances for these two tasks are built based on an existing mathematical instruction datasets with reasoning steps.
Subsequently, LLMs undergo multi-task fine-tuning with an existing mathematical instruction datasets and the newly created datasets to enhance the understanding and execution of instructions. 
To validate the influence of the proposed strategy, comprehensive experiments are conducted on multiple mathematical reasoning tasks in various forms and domains. 
The experimental results indicate that this strategy improve the reasoning abilities and domain generalization of the model.

Overall, this paper makes the following contributions:
(1) A dual instruction tuning strategy is proposed to meticulously model mathematical reasoning from both forward and reverse directions thereby improving the reasoning abilities and domain generalization.
(2) New training instances are created for these two dual tasks, which can be utilized to enhance model's abilities of understanding and execution instructions.
(3) Experimental results demonstrate the effectiveness and domain generalization of the model across various datasets.

\section{Method}
The dual instruction tuning strategy (Section~\ref{subsec: tasks}) involves introducing the Intermediate Reasoning State Prediction task (IRSP) and the Instruction Reconstruction task (IR). 
Two examples to illustrate the process of data construction for these two tasks are shown in Figure~\ref{fig:task} in Appendix.
Subsequently,  LLMs are fine-tuned in a multi-task manner using data consisted of existing mathematical instructions and the two dual tasks(Section~\ref{subsec: train}).

\subsection{Dual Instruction Tuning Strategy}
\label{subsec: tasks}

\paragraph{Intermediate Reasoning State Prediction (IRSP)}
A chain of thought typically comprises a series of intermediate reasoning steps, and inaccuracies in any step directly influence subsequent predictions and, consequently, the final answer prediction. 
This study, therefore, enhances the supervision of intermediate reasoning states to improve the model's understanding and execution of instructions.
The learning objective for thought generation involves the "forward reasoning" process from the instruction space to the thought space.
Concretely, for a $\langle$instruction, CoT\&PoT$\rangle$ pair, we randomly mask certain intermediate reasoning steps in the chain of thought with a \textit{<MASK>} tag.  
Subsequently, as shown in Figure~\ref{fig:task} in Appendix, the model is prompted to predict intermediate reasoning states based on instructions and partially revealed thoughts.
Based on an existing instruction dataset MathInstruct~\cite{DBLP:journals/corr/abs-2309-05653}, we construct the IRSP task data using this approach and incorporate it to the original training set.

\paragraph{Instruction Reconstruction (IR)}
 Drawing inspiration from the concept of dual learning~\cite{DBLP:conf/nips/HeXQWYLM16}, this study introduces the IR task as a learning objective for the "reverse reasoning" process from the thought chain space to the instruction space. 
This addition aims to enhance the model's capability in understanding and executing instructions.
The IR task entails randomly masking certain conditions and questions in the instruction using \textit{<MASK>}.
Subsequently, the model is prompted to reconstruct the masked portions based on thoughts and partially uncovered instructions. 
This study generates IR task data using the $\langle$instruction, CoT\& PoT$\rangle$ data from the MathInstruct dataset~\cite{DBLP:journals/corr/abs-2309-05653} and the constructed data is added to the original training set. 
By combining "forward" and "reverse" reasoning, the model can better learn the mapping relationship between the instruction space and the thought space.

\subsection{Training and Inference}
\label{subsec: train}
Following the MAmmoTH method~\cite{DBLP:journals/corr/abs-2309-05653}, this study adopts a hybrid instruction tuning strategy, including both CoT and PoT data, to absorb the advantages of these two forms of thoughts in the abstract reasoning scenarios and computation precision, respectively.
During instruction tuning, the model input comprises three parts: system instructions, instructions, and target response. 
Specifically, for each training instance, the following system instruction is added to prompt the model to understand instructions and generate response in forms of CoT or PoT.

\texttt{Below is an instruction that describes a task. Write a response that appropriately completes the request.}

\texttt{\#\#\# Human: \{Instruction\}}

\texttt{\#\#\# Assistant: \{Response\}}

\begin{figure}[t]
	\centering
	\includegraphics[width=3.05in]{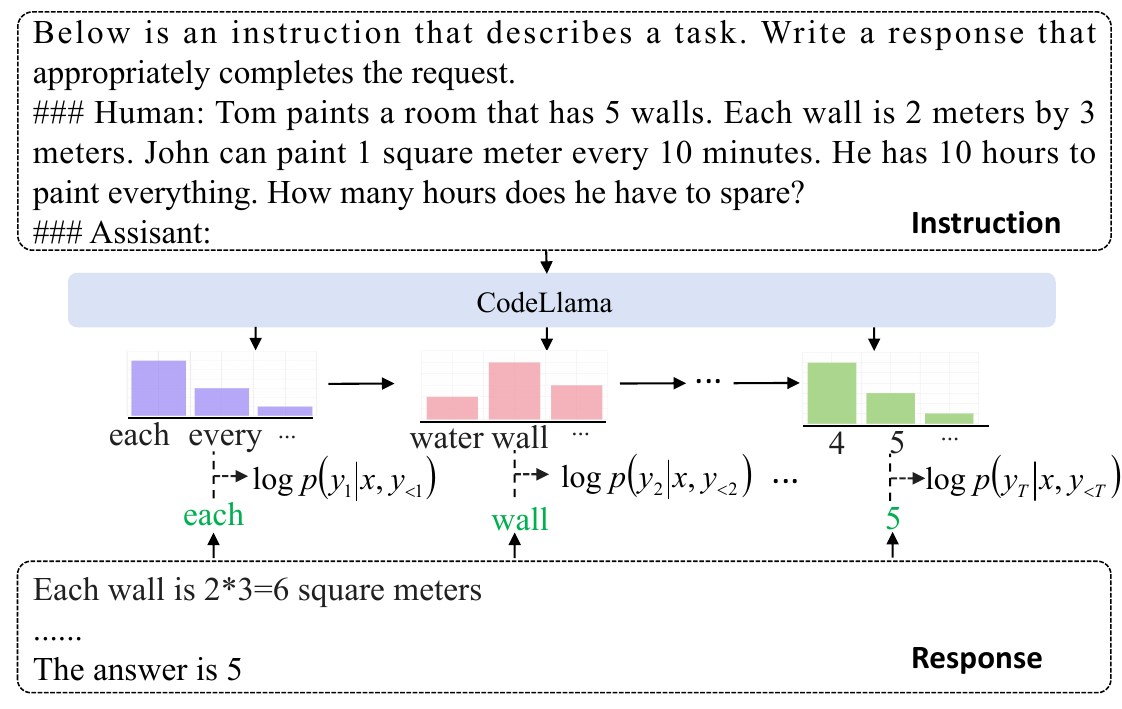}
	\caption{Illustration of instruction tuning process based on the CodeLlama model }
    \label{fig:codellam_model}
\end{figure}
Here, \texttt{\{Instruction\}} refers to user instruction, and \texttt{\{Response\}} represents the target response.
As illustrated in Figure \ref{fig:codellam_model}, during instruction tuning, the model calculates the probability distribution $p(y_t|x, y_{<t};\theta)$ for the $t$-th step vocabulary based on the instruction input $x$ and the response sequence of the previous $t-1$ steps. 
This study investigate the CodeLlama~\cite{DBLP:journals/corr/abs-2308-12950} model as backbone, which belongs to the causal decoder structure, and it takes both the instruction and the target response as inputs but only calculates the loss $\mathcal{L}$ for the target response:

\begin{equation}
\begin{aligned}
\mathcal{L} &= -{\frac{1}{T}\sum_{t=1}^{T}{\log{p(y_t|x; y_{<t}; \theta)}}}.
\end{aligned}
\end{equation}
Here, $\theta$ represents the model parameters, $y_t$ represents the gold output at the $t$-th step, $x$ represents the instruction to the model, and $T$ represents the decoding steps.
When inference, an additional prompt \texttt{Let's write a program.} is added to the \texttt{{Instruction}} thereby prompt the model generates a program of thought.
If the currently generated program of thought cannot produce a result, the model continues to generate a chain of thought. 
For multiple-choice questions, if the generated answer contains options, the options are directly output as the result. 
Otherwise, the model is prompted to choose the option closest to the current answer as the output result using the following prompt.
\texttt{Please find the closest option to {answer}. The options are {options}.}

\begin{table*}[t]
	\centering
	\caption{Performance of baselines and our models on the mathematical reasoning tasks}
	\resizebox{1.0\textwidth}{!}{
	\begin{tabular}{lcccccccc}
		\toprule
		\multirow{2}[4]{*}{Methods} & \multirow{2}[4]{*}{Base} & \multirow{2}[4]{*}{Math-SFT?} & \multicolumn{4}{c}{In-Domain} & \multicolumn{2}{c}{Out-of-Domain} \\
		\cmidrule(lr){4-7}    \cmidrule(lr){8-9}
		 & & & GSM8K & MATH & NumGLUE & AQuA & SVAMP & Mathematics \\
		\midrule
		\multicolumn{9}{c}{Closed-Models} \\ 
		\midrule
		﻿GPT-4~\cite{DBLP:journals/corr/abs-2303-08774} &  & unknown & 92.0    & 42.5  & -- & 72.6  & ﻿97.0 & {--} \\
		GPT-4(code-interpreter)~\cite{DBLP:journals/corr/abs-2303-08774} &  & unknown & 97.0    & 69.7  & {--} & {--} &{--} & {--} \\
		PaLM-2-340B~\cite{DBLP:journals/corr/abs-2305-10403} &  & unknown & 80.7  & 34.3  & {--} & 64.1  & {--} & {--} \\
		Claude-2~\cite{DBLP:journals/corr/abs-2212-08073} &  & unknown & 85.2  & 32.5  & {--} & 60.9  & {--} & {--} \\
		Codex(PoT)-175B~\cite{DBLP:journals/corr/abs-2107-03374} &  & unknown & 71.6  & 36.8  & {--} & 54.1  & 85.2  & {--} \\
		\midrule
		\multicolumn{9}{c}{7B-Models} \\ 
		\midrule
		
		﻿Llama-1~\cite{DBLP:journals/corr/abs-2302-13971} & & No    & 10.7  & 2.9   & 24.7  & 22.6  & 24.5  & 6.2 \\
		﻿Llama-2~\cite{DBLP:journals/corr/abs-2307-09288} &  & No    & 14.6  & 2.5   & 29.9  & 30.3  & 34.5  & 6.0 \\
		﻿CodeLlama (PoT)~\cite{DBLP:journals/corr/abs-2308-12950} &  & No    & 25.2  & 13.0    & 26.8  & 24.0    & 49.4  & 21.7 \\
		﻿AQuA-SFT~\cite{DBLP:conf/acl/LingYDB17} & Llama2 & ﻿AQuA & 11.2  & 3.6   & 12.2  & 35.6  & {--} & {--} \\
		﻿Llama-1 RFT~\cite{yuan2023scaling} & Llama1 & ﻿GSM8K & 46.5  & 5.2   & 21.1  & 18.8  & 21.1  & 5.1 \\
		 Baichuan2-7B~\cite{yang2023baichuan}  & {} & {No} & 24.5 & 5.6 & {--} & {--} & {--} & {--}\\
		 Qwen-7B-Chat~\cite{bai2023qwen} & {} & {Yes} & 50.3 & 6.8  & {--} & {--} & {--} & {--} \\
		 DeepSeek-7B-Chat~\cite{deepseekai2024deepseek} & {} & {Yes} &  63.0 & 15.8 & {--} & {--} & {--} & {--} \\
		﻿WizardMath~\cite{DBLP:journals/corr/abs-2308-09583} & Llama2 & ﻿GSM8K+MATH & 54.9  & 10.7  & 36.1  & 26.3  & 36.1  & 9.3 \\
		﻿MAmmoTH~\cite{DBLP:journals/corr/abs-2309-05653} & Llama2 & ﻿MathInstruct & 53.6  & 31.5  & 61.2  & 44.5  & 67.7  & 46.3 \\
		MAmmoTH-coder~\cite{DBLP:journals/corr/abs-2309-05653} & CodeLlama & ﻿MathInstruct & 59.4  & 33.4  & 66.4  & 47.2  & 71.4  & 55.4 \\
		\midrule
		Baseline(Ours)  & CodeLlama & ﻿MathInstruct & 66.5  & 34.5  & 74.7  & 45.3  & 73.9  & 57.8 \\
		Baseline(Ours) + IRSP & CodeLlama & ﻿MathInstruct & \bf 67.2  & 35.0    & 75.4  & \bf 50.8  & \bf 76.8  & 57.0 \\
	    Baseline(Ours) + IR & CodeLlama & ﻿MathInstruct & 66.1  & \bf 36.0    & \bf 76.3  & 50.4  & 75.2  & \bf 58.8 \\
		\midrule
		\multicolumn{9}{c}{13B-Models} \\ 
		\midrule
		﻿Llama-1~\cite{DBLP:journals/corr/abs-2302-13971} &  & No    & 17.8  & 3.9   & 24.8  & 26.0    & 34.7  & 6.9 \\
		﻿Llama-2~\cite{DBLP:journals/corr/abs-2307-09288} &  & No    & 28.7  & 3.9   & 8.8   & 25.1  & 35.1  & 11.5 \\
		﻿CodeLlama (PoT)~\cite{DBLP:journals/corr/abs-2308-12950} &  & No    & 36.1  & 16.4  & 29.2  & 28.7  & 60.0    & 21.3 \\
		﻿Llama-1 RFT~\cite{yuan2023scaling} & Llama1 & ﻿GSM8K & 52.1  & 5.1   & 24.5  & 16.1  & 46.5  & 6.7 \\
		 Baichuan2-13B~\cite{yang2023baichuan} & {} & {No} & 52.8 & 10.1 & {--} & {--} & {--} & {--}\\
	     Qwen-14B-Chat~\cite{bai2023qwen} & {} & {Yes} & 	60.1 & 18.4  & {--} & {--} & {--} & {--}\\
		﻿WizardMath~\cite{DBLP:journals/corr/abs-2308-09583} & Llama2 & ﻿GSM8K+MATH & 63.9  & 14.0    & 40.8  & 21.2  & 51.9  & 14.1 \\
		﻿MAmmoTH~\cite{DBLP:journals/corr/abs-2309-05653} & Llama2 & ﻿MathInstruct & 62.0    & 34.2  & 68.7  & 51.6  & 72.4  & 49.2 \\
		MAmmoTH-coder~\cite{DBLP:journals/corr/abs-2309-05653} & CodeLlama & ﻿MathInstruct & 64.7  & 36.3  & 66.8  & 46.9  & 73.7  & 61.5 \\
		\midrule
		Baseline(Ours)  & CodeLlama & ﻿MathInstruct & 72.2  & 39.0    & 77.7  & 42.9  & 77.6  & 55.6 \\
		Baseline(Ours) + IRSP & CodeLlama & ﻿MathInstruct & 72.2  & 39.6  & \bf 77.7  & 47.2  & 78.6  & 53.4 \\
		Baseline(Ours) + IR & CodeLlama & ﻿MathInstruct & \bf 72.5  & \bf 39.8  & 77.2  & \bf 51.2  & \bf 79.6  & \bf 63.3 \\
		\bottomrule
	\end{tabular}%
}
\vspace{-0.2cm}
\label{table:performance}
\end{table*}%

\section{Experiment}
\paragraph{Experiment Settings and Evaluation Metrics}
we comprehensively evaluate the mathematical reasoning ability of the model on multiple tasks in various domains and task forms.
The dataset details are seen in Table~\ref{table:eval_data-statics} in Appendix.
The experiments primarily rely on the CodeLlama~\cite{DBLP:journals/corr/abs-2308-12950} in 7B and 13B scales, running on 2*8 80G NVIDIA A800 GPUs. 
The hyper-parameter settings for model training and inference are provided in Table~\ref{table:hyper}.
Accuracy is used as the evaluation metric for answer prediction.
\paragraph{Baselines}
Due to the limitation of length, the details of baselines refers to Appendix~\ref{subsec:baselines}.
\paragraph{Results}
\begin{table}[t]
	\centering
	\caption{Performance improvement of proposed method on error samples}
    \resizebox{0.42\textwidth}{!}{
    \begin{tabular}{cc}
        \toprule
         Datasets & Gains\\
        \midrule
        GSM8K~\cite{DBLP:journals/corr/abs-2110-14168} & 26.0\% \\
        Math~\cite{DBLP:conf/nips/HendrycksBKABTS21} & 7.0\%  \\
        NumGLUE~\cite{DBLP:conf/acl/MishraMVSCBK22} & 31.3\% \\
        AQuA~\cite{DBLP:journals/nature/DaviesVBBZTTBBJ21} & 36.3\%\\
        SVAMP~\cite{DBLP:conf/naacl/PatelBG21} &33.9\%\\ 
        Mathematics~\cite{DBLP:journals/nature/DaviesVBBZTTBBJ21} & 21.5\%\\
        \bottomrule
    \end{tabular}%
    }
\label{table:improvement}
\vspace{-0.5cm}
\end{table}%

Based on the experimental results in Table~\ref{table:performance},
we draw the following two conclusions:
(1) The introduction of the IRSP task results in improvements of 7B model on in-domain tasks like GSM8K, MATH, NumGLUE, and AQuA, as well as on the out-of-domain task SVAMP. The 13B model achieves significant improvements in MATH, AQuA, and SVAMP tasks, but not on relatively simpler tasks like GSM8K and NumGLUE, indicating that the IRSP task is more conducive to enhancing the model's parsing ability for challenging instructions.
Moreover, when introducing the IR task, the model shows noticeable improvements on difficult in-domain tasks like MATH and AQuA, as well as on out-of-domain tasks SVAMP and Mathematics. 
(2) The 13B model surpasses Codex~\cite{DBLP:journals/corr/abs-2107-03374}, Claude-2~\cite{DBLP:journals/corr/abs-2212-08073}, and PaLM-2~\cite{DBLP:journals/corr/abs-2305-10403}, LLMs with 175B+ parameters, in performance on the MATH task but still lags behind the GPT-4 comprehensively. 
The main reasons for this lag are considered to be: (a) Our method relies on the CodeLlama model, which lacks extensive pre-training tailored for mathematical problems, resulting in the model not fully acquire the corresponding mathematical knowledge; (b) This study primarily fine-tunes models with parameters of 7B and 13B scales, which are significantly smaller than closed-source models like GPT-4.

We further analyzes the effect of the strategy on errors samples. 
As shown in Table~\ref{table:improvement}, the proposed method improves errors by more than 20\% on most of tasks, while with a 7\% improvement on the MATH task.
The main reason for the improvement is that the proposed strategy enables the model to accurately learn the relationship between each instruction clause and thought, while the MATH task not only examines the model's understanding and execution of instructions but also critically assesses its understanding and application of mathematical knowledge.

\paragraph{Ablation Study}
We delve into each component of the dual instruction tuning strategy. 
As mentioned before, both IRSP and IR tasks, respectively tuning with the CoT/PoT generation tasks contribute to the performance of models.
As depicted in Table~\ref{table:ablation}, instruction tuning employing both the IRSP and IR task demonstrates remarkable performance in tasks such as Math and Mathematics, particularly those involving multi-domain questions. This observation underscores its significant contribution to enhancing the domain generalization.

\begin{table}[t]
	\centering
	\caption{The ablation results of the proposed method}
    \resizebox{0.49\textwidth}{!}{
    \begin{tabular}{lcccccc}
        \toprule
         Methods & GSM8K & MATH & NumGLUE & AQuA & SVAMP & Mathematics \\ \toprule
		Baseline(Ours)  & 66.5  & 34.5  & 74.7  & 45.3  & 73.9  & 57.8 \\
		~~+ IRSP &  67.2  & 35.0    & 75.4  & 50.8  & 76.8  & 57.0 \\
	    ~~+ IR & 66.1  &  36.0    & 76.3  & 50.4  & 75.2  &  58.8 \\
        ~~+ IRSP + IR & 66.2   & 36.0  &	75.5  &	51.2 &  75.5 & 60.4 \\
        \bottomrule
    \end{tabular}%
    }
\label{table:ablation}
\vspace{-0.4cm}
\end{table}%

\section{Conclusion}
This work propose a dual instruction tuning strategy to alleviate the quality issue of the generated chain of thought with LLMs.
Specifically, this strategy introduce two auxiliary tasks, including Intermediate Reasoning State Prediction and Instruction Reconstruction task, which meticulously models mathematical reasoning from both forward and reverse directions.
Subsequently, additional training data for this two task are constructed and utilized to train LLMs in a multi-task learning manner. 
Experiments verified the proposed method contributes to the mathematical reasoning ability and domain generalization of the model.

\section*{Limitations}
This study aims to enhance the comprehension and execution of instructions, thereby improving the reasoning abilities and domain generalization of LLMs for mathematical reasoning. However, it is important to note that this approach has its limitations when it comes to addressing mathematical reasoning problems that demand a higher level of mathematical knowledge. 
For such complex problems, a more effective solution may involve pre-training the LLMs on extensive training data that incorporates a broad range of mathematical concepts and principles.

\bibliography{anthology}

\begin{thebibliography}{42}
\expandafter\ifx\csname natexlab\endcsname\relax\def\natexlab#1{#1}\fi

\bibitem[{Amini et~al.(2019)Amini, Gabriel, Lin, Koncel{-}Kedziorski, Choi, and Hajishirzi}]{DBLP:conf/naacl/AminiGLKCH19}
Aida Amini, Saadia Gabriel, Shanchuan Lin, Rik Koncel{-}Kedziorski, Yejin Choi, and Hannaneh Hajishirzi. 2019.
\newblock Mathqa: Towards interpretable math word problem solving with operation-based formalisms.
\newblock In \emph{Proceedings of the 2019 Conference of the North American Chapter of the Association for Computational Linguistics: Human Language Technologies}, pages 2357--2367.

\bibitem[{Anil et~al.(2023)Anil, Dai, Firat, Johnson, Lepikhin, Passos, Shakeri, Taropa, Bailey, Chen, Chu, Clark, Shafey, Huang, Meier{-}Hellstern, Mishra, Moreira, Omernick, Robinson, Ruder, Tay, Xiao, Xu, Zhang, {\'{A}}brego, Ahn, Austin, Barham, Botha, Bradbury, Brahma, Brooks, Catasta, Cheng, Cherry, Choquette{-}Choo, Chowdhery, Crepy, Dave, Dehghani, Dev, Devlin, D{\'{\i}}az, Du, Dyer, Feinberg, Feng, Fienber, Freitag, Garcia, Gehrmann, Gonzalez, and et~al.}]{DBLP:journals/corr/abs-2305-10403}
Rohan Anil, Andrew~M. Dai, Orhan Firat, Melvin Johnson, Dmitry Lepikhin, Alexandre Passos, Siamak Shakeri, Emanuel Taropa, Paige Bailey, Zhifeng Chen, Eric Chu, Jonathan~H. Clark, Laurent~El Shafey, Yanping Huang, Kathy Meier{-}Hellstern, Gaurav Mishra, Erica Moreira, Mark Omernick, Kevin Robinson, Sebastian Ruder, Yi~Tay, Kefan Xiao, Yuanzhong Xu, Yujing Zhang, Gustavo~Hern{\'{a}}ndez {\'{A}}brego, Junwhan Ahn, Jacob Austin, Paul Barham, Jan~A. Botha, James Bradbury, Siddhartha Brahma, Kevin Brooks, Michele Catasta, Yong Cheng, Colin Cherry, Christopher~A. Choquette{-}Choo, Aakanksha Chowdhery, Cl{\'{e}}ment Crepy, Shachi Dave, Mostafa Dehghani, Sunipa Dev, Jacob Devlin, Mark D{\'{\i}}az, Nan Du, Ethan Dyer, Vladimir Feinberg, Fangxiaoyu Feng, Vlad Fienber, Markus Freitag, Xavier Garcia, Sebastian Gehrmann, Lucas Gonzalez, and et~al. 2023.
\newblock Palm 2 technical report.
\newblock \emph{arXiv preprint arXiv:2305.10403}.

\bibitem[{Bai et~al.(2023)Bai, Bai, Chu, Cui, Dang, Deng, Fan, Ge, Han, Huang et~al.}]{bai2023qwen}
Jinze Bai, Shuai Bai, Yunfei Chu, Zeyu Cui, Kai Dang, Xiaodong Deng, Yang Fan, Wenbin Ge, Yu~Han, Fei Huang, et~al. 2023.
\newblock Qwen technical report.
\newblock \emph{arXiv preprint arXiv:2309.16609}.

\bibitem[{Bai et~al.(2022)Bai, Kadavath, Kundu, Askell, Kernion, Jones, Chen, Goldie, Mirhoseini, McKinnon, Chen, Olsson, Olah, Hernandez, Drain, Ganguli, Li, Tran{-}Johnson, Perez, Kerr, Mueller, Ladish, Landau, Ndousse, Lukosiute, Lovitt, Sellitto, Elhage, Schiefer, Mercado, DasSarma, Lasenby, Larson, Ringer, Johnston, Kravec, Showk, Fort, Lanham, Telleen{-}Lawton, Conerly, Henighan, Hume, Bowman, Hatfield{-}Dodds, Mann, Amodei, Joseph, McCandlish, Brown, and Kaplan}]{DBLP:journals/corr/abs-2212-08073}
Yuntao Bai, Saurav Kadavath, Sandipan Kundu, Amanda Askell, Jackson Kernion, Andy Jones, Anna Chen, Anna Goldie, Azalia Mirhoseini, Cameron McKinnon, Carol Chen, Catherine Olsson, Christopher Olah, Danny Hernandez, Dawn Drain, Deep Ganguli, Dustin Li, Eli Tran{-}Johnson, Ethan Perez, Jamie Kerr, Jared Mueller, Jeffrey Ladish, Joshua Landau, Kamal Ndousse, Kamile Lukosiute, Liane Lovitt, Michael Sellitto, Nelson Elhage, Nicholas Schiefer, Noem{\'{\i}} Mercado, Nova DasSarma, Robert Lasenby, Robin Larson, Sam Ringer, Scott Johnston, Shauna Kravec, Sheer~El Showk, Stanislav Fort, Tamera Lanham, Timothy Telleen{-}Lawton, Tom Conerly, Tom Henighan, Tristan Hume, Samuel~R. Bowman, Zac Hatfield{-}Dodds, Ben Mann, Dario Amodei, Nicholas Joseph, Sam McCandlish, Tom Brown, and Jared Kaplan. 2022.
\newblock Constitutional {AI:} harmlessness from {AI} feedback.
\newblock \emph{arXiv preprint arXiv:2212.08073}.

\bibitem[{Besta et~al.(2023)Besta, Blach, Kubicek, Gerstenberger, Gianinazzi, Gajda, Lehmann, Podstawski, Niewiadomski, Nyczyk, and Hoefler}]{DBLP:journals/corr/abs-2308-09687}
Maciej Besta, Nils Blach, Ales Kubicek, Robert Gerstenberger, Lukas Gianinazzi, Joanna Gajda, Tomasz Lehmann, Michal Podstawski, Hubert Niewiadomski, Piotr Nyczyk, and Torsten Hoefler. 2023.
\newblock Graph of thoughts: Solving elaborate problems with large language models.
\newblock \emph{arXiv preprint arXiv:2308.09687}.

\bibitem[{Chen et~al.(2021)Chen, Tworek, Jun, Yuan, de~Oliveira~Pinto, Kaplan, Edwards, Burda, Joseph, Brockman, Ray, Puri, Krueger, Petrov, Khlaaf, Sastry, Mishkin, Chan, Gray, Ryder, Pavlov, Power, Kaiser, Bavarian, Winter, Tillet, Such, Cummings, Plappert, Chantzis, Barnes, Herbert{-}Voss, Guss, Nichol, Paino, Tezak, Tang, Babuschkin, Balaji, Jain, Saunders, Hesse, Carr, Leike, Achiam, Misra, Morikawa, Radford, Knight, Brundage, Murati, Mayer, Welinder, McGrew, Amodei, McCandlish, Sutskever, and Zaremba}]{DBLP:journals/corr/abs-2107-03374}
Mark Chen, Jerry Tworek, Heewoo Jun, Qiming Yuan, Henrique~Pond{\'{e}} de~Oliveira~Pinto, Jared Kaplan, Harrison Edwards, Yuri Burda, Nicholas Joseph, Greg Brockman, Alex Ray, Raul Puri, Gretchen Krueger, Michael Petrov, Heidy Khlaaf, Girish Sastry, Pamela Mishkin, Brooke Chan, Scott Gray, Nick Ryder, Mikhail Pavlov, Alethea Power, Lukasz Kaiser, Mohammad Bavarian, Clemens Winter, Philippe Tillet, Felipe~Petroski Such, Dave Cummings, Matthias Plappert, Fotios Chantzis, Elizabeth Barnes, Ariel Herbert{-}Voss, William~Hebgen Guss, Alex Nichol, Alex Paino, Nikolas Tezak, Jie Tang, Igor Babuschkin, Suchir Balaji, Shantanu Jain, William Saunders, Christopher Hesse, Andrew~N. Carr, Jan Leike, Joshua Achiam, Vedant Misra, Evan Morikawa, Alec Radford, Matthew Knight, Miles Brundage, Mira Murati, Katie Mayer, Peter Welinder, Bob McGrew, Dario Amodei, Sam McCandlish, Ilya Sutskever, and Wojciech Zaremba. 2021.
\newblock Evaluating large language models trained on code.
\newblock \emph{arXiv preprint arXiv:2107.03374}.

\bibitem[{Chen(2023)}]{DBLP:conf/eacl/Chen23}
Wenhu Chen. 2023.
\newblock Large language models are few(1)-shot table reasoners.
\newblock In \emph{Findings of the Association for Computational Linguistics}, pages 1090--1100.

\bibitem[{Chen et~al.(2022)Chen, Ma, Wang, and Cohen}]{DBLP:journals/corr/abs-2211-12588}
Wenhu Chen, Xueguang Ma, Xinyi Wang, and William~W. Cohen. 2022.
\newblock Program of thoughts prompting: Disentangling computation from reasoning for numerical reasoning tasks.
\newblock \emph{arXiv preprint arXiv:2211.12588}.

\bibitem[{Chen et~al.(2023)Chen, Yin, Ku, Lu, Wan, Ma, Xu, Wang, and Xia}]{DBLP:journals/corr/abs-2305-12524}
Wenhu Chen, Ming Yin, Max Ku, Pan Lu, Yixin Wan, Xueguang Ma, Jianyu Xu, Xinyi Wang, and Tony Xia. 2023.
\newblock Theoremqa: {A} theorem-driven question answering dataset.
\newblock \emph{arXiv preprint arXiv:2305.12524}.

\bibitem[{Cobbe et~al.(2021)Cobbe, Kosaraju, Bavarian, Chen, Jun, Kaiser, Plappert, Tworek, Hilton, Nakano, Hesse, and Schulman}]{DBLP:journals/corr/abs-2110-14168}
Karl Cobbe, Vineet Kosaraju, Mohammad Bavarian, Mark Chen, Heewoo Jun, Lukasz Kaiser, Matthias Plappert, Jerry Tworek, Jacob Hilton, Reiichiro Nakano, Christopher Hesse, and John Schulman. 2021.
\newblock \href {https://arxiv.org/abs/2110.14168} {Training verifiers to solve math word problems}.
\newblock \emph{CoRR}, abs/2110.14168.

\bibitem[{Davies et~al.(2021)Davies, Velickovic, Buesing, Blackwell, Zheng, Tomasev, Tanburn, Battaglia, Blundell, Juh{\'{a}}sz, Lackenby, Williamson, Hassabis, and Kohli}]{DBLP:journals/nature/DaviesVBBZTTBBJ21}
Alex Davies, Petar Velickovic, Lars Buesing, Sam Blackwell, Daniel Zheng, Nenad Tomasev, Richard Tanburn, Peter~W. Battaglia, Charles Blundell, Andr{\'{a}}s Juh{\'{a}}sz, Marc Lackenby, Geordie Williamson, Demis Hassabis, and Pushmeet Kohli. 2021.
\newblock Advancing mathematics by guiding human intuition with {AI}.
\newblock \emph{Nature}, 600(7887):70--74.

\bibitem[{DeepSeek-AI(2024)}]{deepseekai2024deepseek}
DeepSeek-AI. 2024.
\newblock Deepseek llm: Scaling open-source language models with longtermism.
\newblock \emph{arXiv preprint arXiv:2309.16609}.

\bibitem[{Diao et~al.(2023)Diao, Wang, Lin, and Zhang}]{DBLP:journals/corr/abs-2302-12246}
Shizhe Diao, Pengcheng Wang, Yong Lin, and Tong Zhang. 2023.
\newblock Active prompting with chain-of-thought for large language models.
\newblock \emph{arXiv preprint arXiv:2302.12246}.

\bibitem[{Du et~al.(2022)Du, Qian, Liu, Ding, Qiu, Yang, and Tang}]{DBLP:conf/acl/DuQLDQY022}
Zhengxiao Du, Yujie Qian, Xiao Liu, Ming Ding, Jiezhong Qiu, Zhilin Yang, and Jie Tang. 2022.
\newblock {GLM:} general language model pretraining with autoregressive blank infilling.
\newblock In \emph{Proceedings of the 60th Annual Meeting of the Association for Computational Linguistics}, pages 320--335.

\bibitem[{Gao et~al.(2023)Gao, Madaan, Zhou, Alon, Liu, Yang, Callan, and Neubig}]{DBLP:journals/corr/abs-2211-10435}
Luyu Gao, Aman Madaan, Shuyan Zhou, Uri Alon, Pengfei Liu, Yiming Yang, Jamie Callan, and Graham Neubig. 2023.
\newblock {PAL:} program-aided language models.
\newblock In \emph{International Conference on Machine Learning}, pages 10764--10799.

\bibitem[{He et~al.(2016)He, Xia, Qin, Wang, Yu, Liu, and Ma}]{DBLP:conf/nips/HeXQWYLM16}
Di~He, Yingce Xia, Tao Qin, Liwei Wang, Nenghai Yu, Tie{-}Yan Liu, and Wei{-}Ying Ma. 2016.
\newblock Dual learning for machine translation.
\newblock In \emph{Advances in Neural Information Processing Systems 29}, pages 820--828.

\bibitem[{Hendrycks et~al.(2021)Hendrycks, Burns, Kadavath, Arora, Basart, Tang, Song, and Steinhardt}]{DBLP:conf/nips/HendrycksBKABTS21}
Dan Hendrycks, Collin Burns, Saurav Kadavath, Akul Arora, Steven Basart, Eric Tang, Dawn Song, and Jacob Steinhardt. 2021.
\newblock Measuring mathematical problem solving with the {MATH} dataset.
\newblock In \emph{Proceedings of the Neural Information Processing Systems Track on Datasets and Benchmarks 1, NeurIPS Datasets and Benchmarks 2021, December 2021, virtual}.

\bibitem[{Li et~al.(2023)Li, Hammoud, Itani, Khizbullin, and Ghanem}]{DBLP:journals/corr/abs-2303-17760}
Guohao Li, Hasan Abed Al~Kader Hammoud, Hani Itani, Dmitrii Khizbullin, and Bernard Ghanem. 2023.
\newblock {CAMEL:} communicative agents for "mind" exploration of large scale language model society.
\newblock \emph{arXiv preprint arXiv:2303.17760}.

\bibitem[{Ling et~al.(2017)Ling, Yogatama, Dyer, and Blunsom}]{DBLP:conf/acl/LingYDB17}
Wang Ling, Dani Yogatama, Chris Dyer, and Phil Blunsom. 2017.
\newblock Program induction by rationale generation: Learning to solve and explain algebraic word problems.
\newblock In \emph{Proceedings of the 55th Annual Meeting of the Association for Computational Linguistics, {ACL} 2017, Vancouver, Canada, July 30 - August 4, Volume 1: Long Papers}, pages 158--167. Association for Computational Linguistics.

\bibitem[{Luo et~al.(2023{\natexlab{a}})Luo, Sun, Xu, Zhao, Lou, Tao, Geng, Lin, Chen, and Zhang}]{DBLP:journals/corr/abs-2308-09583}
Haipeng Luo, Qingfeng Sun, Can Xu, Pu~Zhao, Jianguang Lou, Chongyang Tao, Xiubo Geng, Qingwei Lin, Shifeng Chen, and Dongmei Zhang. 2023{\natexlab{a}}.
\newblock Wizardmath: Empowering mathematical reasoning for large language models via reinforced evol-instruct.
\newblock \emph{arXiv preprint arXiv:2308.09583}.

\bibitem[{Luo et~al.(2023{\natexlab{b}})Luo, Xu, Zhao, Sun, Geng, Hu, Tao, Ma, Lin, and Jiang}]{DBLP:journals/corr/abs-2306-08568}
Ziyang Luo, Can Xu, Pu~Zhao, Qingfeng Sun, Xiubo Geng, Wenxiang Hu, Chongyang Tao, Jing Ma, Qingwei Lin, and Daxin Jiang. 2023{\natexlab{b}}.
\newblock Wizardcoder: Empowering code large language models with evol-instruct.
\newblock \emph{arXiv preprint arXiv:2306.08568}.

\bibitem[{Mishra et~al.(2022{\natexlab{a}})Mishra, Finlayson, Lu, Tang, Welleck, Baral, Rajpurohit, Tafjord, Sabharwal, Clark, and Kalyan}]{DBLP:conf/emnlp/MishraFLTWBRTSC22}
Swaroop Mishra, Matthew Finlayson, Pan Lu, Leonard Tang, Sean Welleck, Chitta Baral, Tanmay Rajpurohit, Oyvind Tafjord, Ashish Sabharwal, Peter Clark, and Ashwin Kalyan. 2022{\natexlab{a}}.
\newblock {LILA:} {A} unified benchmark for mathematical reasoning.
\newblock In \emph{Proceedings of the 2022 Conference on Empirical Methods in Natural Language Processing}, pages 5807--5832.

\bibitem[{Mishra et~al.(2022{\natexlab{b}})Mishra, Mitra, Varshney, Sachdeva, Clark, Baral, and Kalyan}]{DBLP:conf/acl/MishraMVSCBK22}
Swaroop Mishra, Arindam Mitra, Neeraj Varshney, Bhavdeep~Singh Sachdeva, Peter Clark, Chitta Baral, and Ashwin Kalyan. 2022{\natexlab{b}}.
\newblock Numglue: {A} suite of fundamental yet challenging mathematical reasoning tasks.
\newblock In \emph{Proceedings of the 60th Annual Meeting of the Association for Computational Linguistics}, pages 3505--3523.

\bibitem[{OpenAI(2023)}]{DBLP:journals/corr/abs-2303-08774}
OpenAI. 2023.
\newblock {GPT-4} technical report.
\newblock \emph{arXiv preprint arXiv:2303.08774}.

\bibitem[{Patel et~al.(2021)Patel, Bhattamishra, and Goyal}]{DBLP:conf/naacl/PatelBG21}
Arkil Patel, Satwik Bhattamishra, and Navin Goyal. 2021.
\newblock Are {NLP} models really able to solve simple math word problems?
\newblock In \emph{Proceedings of the 2021 Conference of the North American Chapter of the Association for Computational Linguistics: Human Language Technologies, {NAACL-HLT} 2021, Online, June 6-11, 2021}, pages 2080--2094. Association for Computational Linguistics.

\bibitem[{Rozi{\`{e}}re et~al.(2023)Rozi{\`{e}}re, Gehring, Gloeckle, Sootla, Gat, Tan, Adi, Liu, Remez, Rapin, Kozhevnikov, Evtimov, Bitton, Bhatt, Canton{-}Ferrer, Grattafiori, Xiong, D{\'{e}}fossez, Copet, Azhar, Touvron, Martin, Usunier, Scialom, and Synnaeve}]{DBLP:journals/corr/abs-2308-12950}
Baptiste Rozi{\`{e}}re, Jonas Gehring, Fabian Gloeckle, Sten Sootla, Itai Gat, Xiaoqing~Ellen Tan, Yossi Adi, Jingyu Liu, Tal Remez, J{\'{e}}r{\'{e}}my Rapin, Artyom Kozhevnikov, Ivan Evtimov, Joanna Bitton, Manish Bhatt, Cristian Canton{-}Ferrer, Aaron Grattafiori, Wenhan Xiong, Alexandre D{\'{e}}fossez, Jade Copet, Faisal Azhar, Hugo Touvron, Louis Martin, Nicolas Usunier, Thomas Scialom, and Gabriel Synnaeve. 2023.
\newblock Code llama: Open foundation models for code.
\newblock \emph{arXiv preprint arXiv:2308.12950}.

\bibitem[{Touvron et~al.(2023{\natexlab{a}})Touvron, Lavril, Izacard, Martinet, Lachaux, Lacroix, Rozi{\`{e}}re, Goyal, Hambro, Azhar, Rodriguez, Joulin, Grave, and Lample}]{DBLP:journals/corr/abs-2302-13971}
Hugo Touvron, Thibaut Lavril, Gautier Izacard, Xavier Martinet, Marie{-}Anne Lachaux, Timoth{\'{e}}e Lacroix, Baptiste Rozi{\`{e}}re, Naman Goyal, Eric Hambro, Faisal Azhar, Aur{\'{e}}lien Rodriguez, Armand Joulin, Edouard Grave, and Guillaume Lample. 2023{\natexlab{a}}.
\newblock Llama: Open and efficient foundation language models.
\newblock \emph{arXiv preprint arXiv:2302.13971}.

\bibitem[{Touvron et~al.(2023{\natexlab{b}})Touvron, Martin, Stone, Albert, Almahairi, Babaei, Bashlykov, Batra, Bhargava, Bhosale, Bikel, Blecher, Canton{-}Ferrer, Chen, Cucurull, Esiobu, Fernandes, Fu, Fu, Fuller, Gao, Goswami, Goyal, Hartshorn, Hosseini, Hou, Inan, Kardas, Kerkez, Khabsa, Kloumann, Korenev, Koura, Lachaux, Lavril, Lee, Liskovich, Lu, Mao, Martinet, Mihaylov, Mishra, Molybog, Nie, Poulton, Reizenstein, Rungta, Saladi, Schelten, Silva, Smith, Subramanian, Tan, Tang, Taylor, Williams, Kuan, Xu, Yan, Zarov, Zhang, Fan, Kambadur, Narang, Rodriguez, Stojnic, Edunov, and Scialom}]{DBLP:journals/corr/abs-2307-09288}
Hugo Touvron, Louis Martin, Kevin Stone, Peter Albert, Amjad Almahairi, Yasmine Babaei, Nikolay Bashlykov, Soumya Batra, Prajjwal Bhargava, Shruti Bhosale, Dan Bikel, Lukas Blecher, Cristian Canton{-}Ferrer, Moya Chen, Guillem Cucurull, David Esiobu, Jude Fernandes, Jeremy Fu, Wenyin Fu, Brian Fuller, Cynthia Gao, Vedanuj Goswami, Naman Goyal, Anthony Hartshorn, Saghar Hosseini, Rui Hou, Hakan Inan, Marcin Kardas, Viktor Kerkez, Madian Khabsa, Isabel Kloumann, Artem Korenev, Punit~Singh Koura, Marie{-}Anne Lachaux, Thibaut Lavril, Jenya Lee, Diana Liskovich, Yinghai Lu, Yuning Mao, Xavier Martinet, Todor Mihaylov, Pushkar Mishra, Igor Molybog, Yixin Nie, Andrew Poulton, Jeremy Reizenstein, Rashi Rungta, Kalyan Saladi, Alan Schelten, Ruan Silva, Eric~Michael Smith, Ranjan Subramanian, Xiaoqing~Ellen Tan, Binh Tang, Ross Taylor, Adina Williams, Jian~Xiang Kuan, Puxin Xu, Zheng Yan, Iliyan Zarov, Yuchen Zhang, Angela Fan, Melanie Kambadur, Sharan Narang, Aur{\'{e}}lien Rodriguez, Robert Stojnic, Sergey Edunov,
  and Thomas Scialom. 2023{\natexlab{b}}.
\newblock Llama 2: Open foundation and fine-tuned chat models.
\newblock \emph{arXiv preprint arXiv:2307.09288}.

\bibitem[{Wang et~al.(2022{\natexlab{a}})Wang, Deng, and Sun}]{DBLP:conf/emnlp/Wang0S22}
Boshi Wang, Xiang Deng, and Huan Sun. 2022{\natexlab{a}}.
\newblock Iteratively prompt pre-trained language models for chain of thought.
\newblock In \emph{Proceedings of the 2022 Conference on Empirical Methods in Natural Language Processing}, pages 2714--2730.

\bibitem[{Wang et~al.(2023{\natexlab{a}})Wang, Ren, Zhou, Lu, Luo, Shi, Zhang, Song, Zhan, and Li}]{DBLP:journals/corr/abs-2310-03731}
Ke~Wang, Houxing Ren, Aojun Zhou, Zimu Lu, Sichun Luo, Weikang Shi, Renrui Zhang, Linqi Song, Mingjie Zhan, and Hongsheng Li. 2023{\natexlab{a}}.
\newblock Mathcoder: Seamless code integration in llms for enhanced mathematical reasoning.
\newblock \emph{arXiv preprint arXiv:2310.03731}.

\bibitem[{Wang et~al.(2023{\natexlab{b}})Wang, Wei, Schuurmans, Le, Chi, and Zhou}]{DBLP:journals/corr/abs-2203-11171}
Xuezhi Wang, Jason Wei, Dale Schuurmans, Quoc~V. Le, Ed~H. Chi, and Denny Zhou. 2023{\natexlab{b}}.
\newblock Self-consistency improves chain of thought reasoning in language models.
\newblock In \emph{The Eleventh International Conference on Learning Representations}.

\bibitem[{Wang et~al.(2022{\natexlab{b}})Wang, Kordi, Mishra, Liu, Smith, Khashabi, and Hajishirzi}]{DBLP:journals/corr/abs-2212-10560}
Yizhong Wang, Yeganeh Kordi, Swaroop Mishra, Alisa Liu, Noah~A. Smith, Daniel Khashabi, and Hannaneh Hajishirzi. 2022{\natexlab{b}}.
\newblock Self-instruct: Aligning language model with self generated instructions.
\newblock \emph{arXiv preprint arXiv:2212.10560}.

\bibitem[{Wei et~al.(2022)Wei, Wang, Schuurmans, Bosma, Ichter, Xia, Chi, Le, and Zhou}]{DBLP:journals/corr/abs-2201-11903}
Jason Wei, Xuezhi Wang, Dale Schuurmans, Maarten Bosma, Brian Ichter, Fei Xia, Ed~H. Chi, Quoc~V. Le, and Denny Zhou. 2022.
\newblock Chain-of-thought prompting elicits reasoning in large language models.
\newblock In \emph{Advances in Neural Information Processing Systems 35}, pages 24824--24837.

\bibitem[{Weng et~al.(2023)Weng, Zhu, He, Liu, and Zhao}]{DBLP:journals/corr/abs-2212-09561}
Yixuan Weng, Minjun Zhu, Shizhu He, Kang Liu, and Jun Zhao. 2023.
\newblock Large language models are reasoners with self-verification.
\newblock In \emph{Findings of the Association for Computational Linguistics: {EMNLP}}, pages 2550--2575.

\bibitem[{Yang et~al.(2023)Yang, Xiao, Wang, Zhang, Bian, Yin, Lv, Pan, Wang, Yan et~al.}]{yang2023baichuan}
Aiyuan Yang, Bin Xiao, Bingning Wang, Borong Zhang, Ce~Bian, Chao Yin, Chenxu Lv, Da~Pan, Dian Wang, Dong Yan, et~al. 2023.
\newblock Baichuan 2: Open large-scale language models.
\newblock \emph{arXiv preprint arXiv:2309.10305}.

\bibitem[{Yao et~al.(2023)Yao, Yu, Zhao, Shafran, Griffiths, Cao, and Narasimhan}]{DBLP:journals/corr/abs-2305-10601}
Shunyu Yao, Dian Yu, Jeffrey Zhao, Izhak Shafran, Thomas~L. Griffiths, Yuan Cao, and Karthik Narasimhan. 2023.
\newblock Tree of thoughts: Deliberate problem solving with large language models.
\newblock \emph{arXiv preprint arXiv:2305.10601}.

\bibitem[{Yu et~al.(2023)Yu, Jiang, Shi, Yu, Liu, Zhang, Kwok, Li, Weller, and Liu}]{DBLP:journals/corr/abs-2309-12284}
Longhui Yu, Weisen Jiang, Han Shi, Jincheng Yu, Zhengying Liu, Yu~Zhang, James~T. Kwok, Zhenguo Li, Adrian Weller, and Weiyang Liu. 2023.
\newblock Metamath: Bootstrap your own mathematical questions for large language models.
\newblock \emph{arXiv preprint arXiv:2309.12284}.

\bibitem[{Yuan et~al.(2023{\natexlab{a}})Yuan, Yuan, Li, Dong, Tan, and Zhou}]{yuan2023scaling}
Zheng Yuan, Hongyi Yuan, Chengpeng Li, Guanting Dong, Chuanqi Tan, and Chang Zhou. 2023{\natexlab{a}}.
\newblock Scaling relationship on learning mathematical reasoning with large language models.
\newblock \emph{arXiv preprint arXiv:2308.01825}.

\bibitem[{Yuan et~al.(2023{\natexlab{b}})Yuan, Yuan, Li, Dong, Tan, and Zhou}]{DBLP:journals/corr/abs-2308-01825}
Zheng Yuan, Hongyi Yuan, Chengpeng Li, Guanting Dong, Chuanqi Tan, and Chang Zhou. 2023{\natexlab{b}}.
\newblock Scaling relationship on learning mathematical reasoning with large language models.
\newblock \emph{arXiv preprint arXiv:2308.01825}.

\bibitem[{Yue et~al.(2023)Yue, Qu, Zhang, Fu, Huang, Sun, Su, and Chen}]{DBLP:journals/corr/abs-2309-05653}
Xiang Yue, Xingwei Qu, Ge~Zhang, Yao Fu, Wenhao Huang, Huan Sun, Yu~Su, and Wenhu Chen. 2023.
\newblock Mammoth: Building math generalist models through hybrid instruction tuning.
\newblock \emph{arXiv preprint arXiv:2309.05653}.

\bibitem[{Zhang et~al.(2023)Zhang, Yang, Yuan, and Yao}]{DBLP:journals/corr/abs-2308-04371}
Yifan Zhang, Jingqin Yang, Yang Yuan, and Andrew~Chi{-}Chih Yao. 2023.
\newblock Cumulative reasoning with large language models.
\newblock \emph{arXiv preprint arXiv:2308.04371}.

\bibitem[{Zhou et~al.(2022)Zhou, Sch{\"{a}}rli, Hou, Wei, Scales, Wang, Schuurmans, Bousquet, Le, and Chi}]{DBLP:journals/corr/abs-2205-10625}
Denny Zhou, Nathanael Sch{\"{a}}rli, Le~Hou, Jason Wei, Nathan Scales, Xuezhi Wang, Dale Schuurmans, Olivier Bousquet, Quoc Le, and Ed~H. Chi. 2022.
\newblock Least-to-most prompting enables complex reasoning in large language models.
\newblock In \emph{The Eleventh International Conference on Learning Representations}.

\end{thebibliography}
\appendix

\section{Appendix}
\label{sec:appendix}
\subsection{Implement Details} 
\paragraph{Training Data and Evaluation Tasks}
The models undergo fine-tuning using the MathInstruct dataset~\cite{DBLP:journals/corr/abs-2309-05653}, which comprises a large-scale collection of hybrid mathematical instructions. Additionally, newly created data for the Intermediate Reasoning State Prediction (IRSP) and Intermediate Reasoning (IR) tasks is incorporated.
Specifically, the MathInstruct dataset includes 18.8w training examples in the format of $\langle$Instruction, CoT$\rangle$ and 7.3w examples in $\langle$Instruction, PoT$\rangle$, totaling 26.2w examples. An overview of the sources and statistical information related to the dataset is provided in Table \ref{table:train-data-statics}.
Furthermore, we assess the effectiveness and domain generalization of the dual instruction tuning strategy across various mathematical reasoning tasks, as detailed in Table~\ref{table:eval_data-statics}.
Concretely, GSM8K\cite{DBLP:journals/corr/abs-2110-14168}, MATH\cite{DBLP:conf/nips/HendrycksBKABTS21}, NumGLUE\cite{DBLP:conf/acl/MishraMVSCBK22}, and AQuA\cite{DBLP:conf/acl/LingYDB17} are selected as in-domain benchmarks. 
Additionally, this study assesses the out-of-domain generalization ability based on SVAMP\cite{DBLP:conf/naacl/PatelBG21} and Mathematics dataset\cite{DBLP:journals/nature/DaviesVBBZTTBBJ21}.
\begin{figure*}[t]
	\centering
	\includegraphics[width=6in]{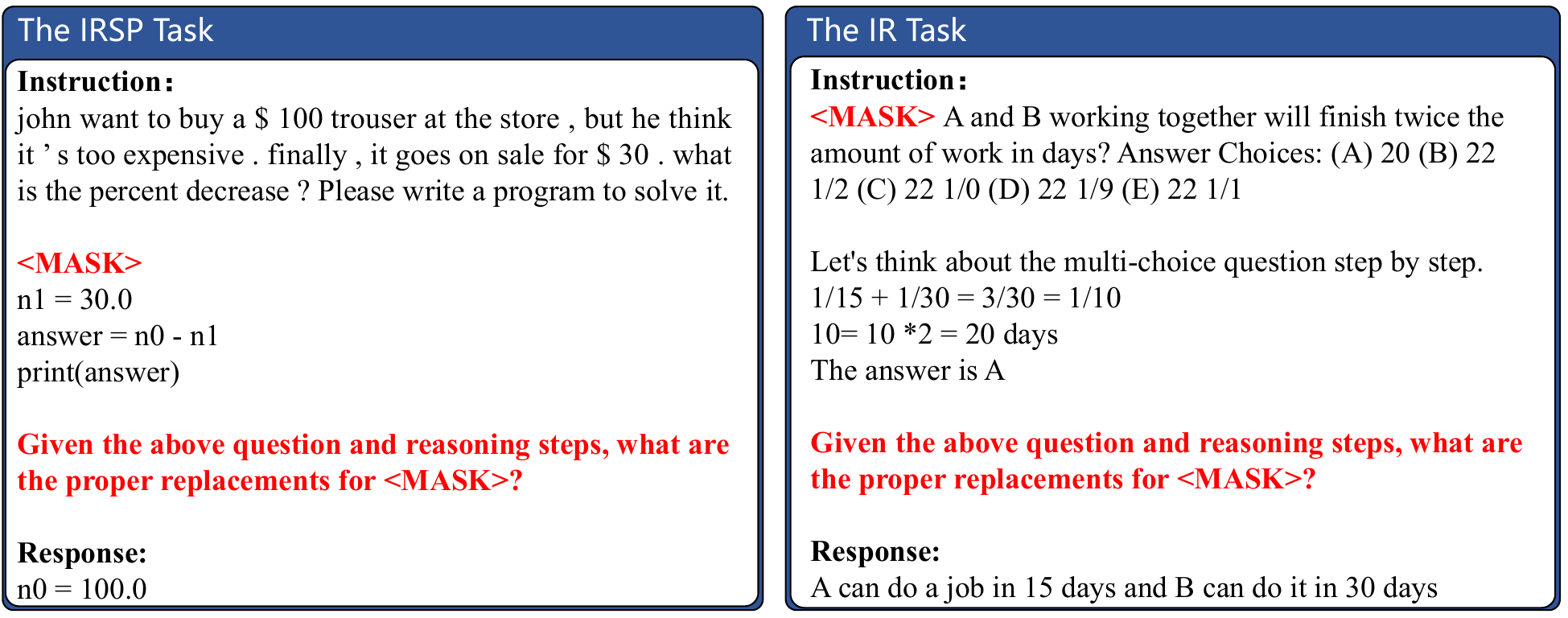}
	\caption{Two examples for the IRSP and IR tasks}
    \label{fig:task}
 
\end{figure*}

\begin{table*}[htbp]
	\centering
	\caption{Mathematical instruction data sources and their statistical information. (1) Pre-Algebra; (2) Inter-Algebra; (3) Algebra; (4) Probability; (5) NumTheory; (6) Calculus; (7) Geometry.}
    \resizebox{0.8\textwidth}{!}{
	\begin{tabular}{lccccc}
		\toprule
		Training Set   & Type   & Annotation    & Size    & Answer Form  & Domains \\
		\midrule
		GSM8K\cite{DBLP:journals/corr/abs-2110-14168}  & CoT   & Human & 7K    &  Open-formed  & (1) \\
		GSM8K-RFT\cite{DBLP:journals/corr/abs-2308-01825} & CoT   & Llama & 28K   &  Open-formed  & (1) \\
		AQuA-RAT\cite{DBLP:conf/acl/LingYDB17} & CoT   & Human & 90K   & Multi-choice  & (2) \\
		MATH\cite{DBLP:conf/nips/HendrycksBKABTS21}  & CoT   & GPT-4 & 7K    &  Open-formed  & (1,2,3,4,5,6,7)\\
		TheoremQA\cite{DBLP:journals/corr/abs-2305-12524}  & CoT   & GPT-4 & 600   &  Open-formed  & (3,4,5,6,7 \\
		Camel-Math\cite{DBLP:journals/corr/abs-2303-17760} & CoT   & GPT-4 & 50K   &  Open-formed  & (3,4,5,6,8) \\
		College-Math & CoT   & GPT-4 & 1.8K  &  Open-formed  & (3) \\
		\midrule
		GSM8K\cite{DBLP:journals/corr/abs-2110-14168} & PoT   & GPT-4 & 14K   &  Open-formed  & (1) \\
		AQuA-RAT\cite{DBLP:conf/acl/LingYDB17} & PoT   & GPT-4 & 9.7K  & Multi-choice  & (2) \\
		MATH\cite{DBLP:conf/nips/HendrycksBKABTS21} & PoT   & GPT-4 & 7K    &  Open-formed  & (1,2,3,4) \\
		TheoremQA\cite{DBLP:journals/corr/abs-2305-12524} & PoT   & GPT-4 & 700   &  Open-formed  & (3,4,5,6,7) \\
		MathQA\cite{DBLP:conf/naacl/AminiGLKCH19}  & PoT   & Human & 25K   &  Open-formed  & (2) \\
		NumGLUE\cite{DBLP:conf/emnlp/MishraFLTWBRTSC22}  & PoT   & Human & 13K   &  Open-formed  & (1) \\
		\bottomrule
	\end{tabular}
 }
 \label{table:train-data-statics}
\end{table*}

\begin{table}[h]
	\centering
	\caption{All the evaluation benchmarks for mathematical reasoning problem and their statics. (1) Pre-Algebra; (2) Inter-Algebra; (3) Algebra; (4) Probability; (5) NumTheory; (6) Calculus; (7) Geometry. }
    \resizebox{0.48\textwidth}{!}{
    \begin{tabular}{cccc}
        \toprule
        Dataset & Size & Answer Form &  Domains \\ \midrule
        GSM8K & 1319 &  Open-formed &  (1)\\
        MATH & 5000 &  Open-formed &  (1,2,3,4,5,6,7)\\
        NumGLUE & 1042 &  Open-formed & (1)\\
        AQuA & 254 &  Multi-choice & (2)\\
        SVAMP & 1000 &  Open-formed & (1) \\
        Mathematics & 1000 &  Open-formed & (1,2,5,6)\\
        \bottomrule
    \end{tabular}
    }
 \label{table:eval_data-statics}
\end{table}

\paragraph{Parameter Settings}
This study primarily conducts experiments based on the CodeLlama models with 7B and 13B parameters, with the hyper-parameter settings for model training and inference outlined in Table~\ref{table:hyper}.

\begin{table}[t]
	\small
	\centering
	\caption{The Hyper-parameters for training and inference on the numerical reasoning tasks}
		\setlength{\tabcolsep}{1mm}{
			\begin{tabular}{lcc}
				\toprule
				Parameters &   7B-Model & 13B-Model \\
				\midrule
				Learning Rate & 2e-5 & 2e-5 \\
				Batch Size & 128 & 128 \\
				Weight Decay & 0.01 & 0.01 \\
				Gradient Clip & 1.0 & 1.0\\
				Warmup Ratio& 0.03 & 0.03 \\
				Warmup Type & Cosine & Cosine\\
				Context Length & 2048 & 2048 \\
				Epochs & 3 & 3 \\
				Max Decoding Length & 1500 & 1500 \\
				\bottomrule
			\end{tabular}
		} 
\label{table:hyper}
\end{table}

\subsection{Details of Baselines}
\label{subsec:baselines}
This study provides a comprehensive comparison of the performance of the mainstream LLMs on mathematical reasoning tasks. 
The results of baseline models come from prior public work \cite{DBLP:journals/corr/abs-2309-05653,yang2023baichuan,bai2023qwen,deepseekai2024deepseek}, which are categorized into the following types:

\textbf{Closed-source models}: This category primarily considers five mainstream closed-source LLMs, namely GPT-4 \cite{DBLP:journals/corr/abs-2303-08774}, GPT-4 (Code Interpreter), PaLM-2 Unicorn~\cite{DBLP:journals/corr/abs-2305-10403}, Claude-2~\cite{DBLP:journals/corr/abs-2212-08073}, and Codex~\cite{DBLP:journals/corr/abs-2107-03374}. GPT-4, PaLM-2 Unicorn, and Claude-2 use the CoT approach for reasoning, while GPT-4 (Code Interpreter) and Codex use the PoT approach to obtain final answers. 

\textbf{Open-source models}: This category mainly considers Llama-1~\cite{DBLP:journals/corr/abs-2302-13971}, Llama-2, and CodeLlama~\cite{DBLP:journals/corr/abs-2308-12950} models. Llama-1 and Llama-2 use the CoT approach to obtain problem answers, while CodeLlama utilizes the PoT approach for reasoning.  
In addition, this study compares models such as Baichuan2-7B~\cite{yang2023baichuan}, Baichuan2-13B, Qwen-7B-Chat~\cite{bai2023qwen}, Qwen-13B-Chat, and DeepSeek-7B-Chat~\cite{deepseekai2024deepseek}. 
Models with "Chat" as suffix indicate they have undergone instruction tuning.

\textbf{Mathematical instruction fine-tuned models (Math-SFT)}: This category mainly considers AQuA-SFT~\cite{DBLP:conf/acl/LingYDB17}, Llama-1 RFT~\cite{yuan2023scaling}, WizardMath~\cite{DBLP:journals/corr/abs-2308-09583}, MAmmoTH~\cite{DBLP:journals/corr/abs-2309-05653}, and the MAmmoTH-coder model. 
Specifically, the AQuA-SFT~\cite{DBLP:conf/acl/LingYDB17} model is fine-tuned based on the Llama model on the AQuA dataset; 
Llama-1 RFT~\cite{yuan2023scaling} is fine-tuned on the GSM8K training set using a rejection sampling strategy for the Llama-1 model. 
The WizardMath model undergoes initial fine-tuning on the base model, followed by the collection of a large amount of mathematical instruction data through an instruction evolution process. 
It then utilizes a reward model and process supervision model for reinforcement learning training. 
\citet{DBLP:journals/corr/abs-2309-05653} collected multiple domain, task formats, and different difficulty level public mathematical datasets and generated a substantial amount of CoT and PoT-formulated response data using GPT-4.
Based on these $\langle$ Instruction, CoT\&PoT$\rangle$, the Llama and CodeLlama models are used as base models for a hybird CoT and PoT instruction tuning process, resulting in the MAmmoTH model.

The Baseline(Ours) models in Table \ref{table:performance} represent the reproduction of MAmmoTH-Coder results by instruction tuning with the MathInstruct dataset in this study. 
The distinction between the Baseline(Ours) models and the MAmmoTH-Coder model lies in system instructions and some varying hyper-parameters.
In most baseline methods, CoT is used to obtain answers to problems, while code models generate PoT-formulated thoughts and execute them using a code interpreter to obtain problem answers. 
The result of MAmmoTH and MAmmoTH-Coder are derived under a zero-shot setting, whereas other baseline model results are under zero-shot and few-shot conditions. 
For ease of comparison regarding the improvement in performance over the MAmmoTH models, the results of our proposed method are reported under a zero-shot condition. 
In the case of multiple-choice questions, if the predicted answer is not directly one of the final options, the model is guided to select the option closest to the predicted value through prompting.

\subsection{Analysis and Discussion}
\label{subsec:analysis}
\begin{figure*}[t]
	\centering
	\includegraphics[width=6.5in]{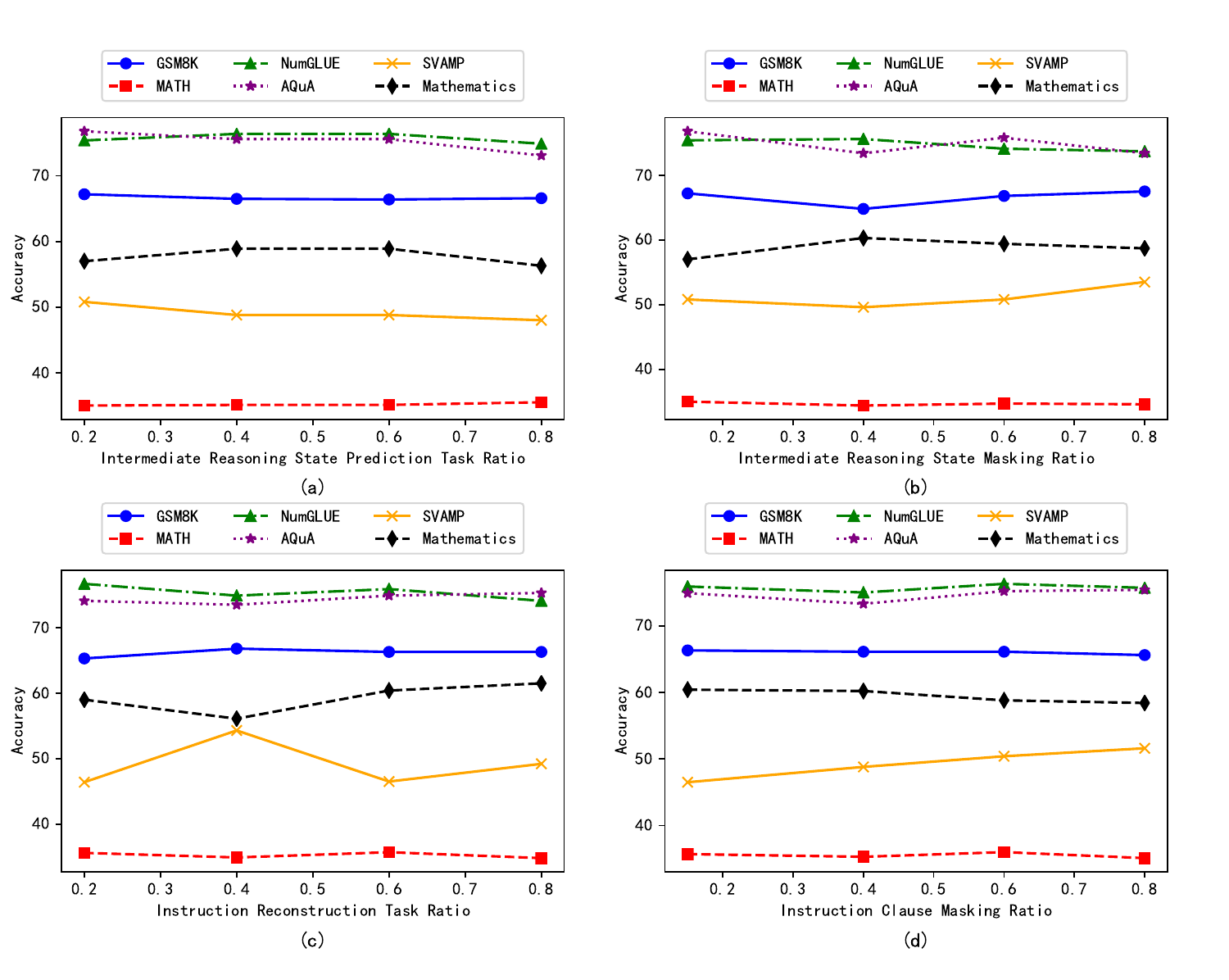}
	\caption{The proportion of data for Intermediate Reasoning State Prediction and Instruction Reconstruction tasks, as well as the masking ratio of reasoning states and instruction}
 \label{fig:ratio_data}
\end{figure*}
In this section, various proportion of data of IRSP/IR task and corresponding mask ratio are investigated to achieve the optimal results.

\paragraph{Intermediate Reasoning State Prediction Task Ratio}
We investigate the influence of the composition ratio of data for IRSP tasks and thought generation on the performance of models using CodeLlama-7B as the base model.
We consider introducing 20\%, 40\%, 60\%, and 80\% proportions of IRSP task data, and 15\% of the intermediate reasoning states in thoughts are randomly masked.
As shown in subfigure (a) of Figure \ref{fig:ratio_data}, the best performance on evaluation tasks like GSM8K, AuQA, and SVAMP is achieved when the task ratio $r_{task}=0.2$, while tasks like NumGLUE, Mathematics, and MATH perform best at $r_{task}=0.4$ or $r_{task}=0.6$. 
Considering both training efficiency and performance, this study adopts a 20\% proportion of IRSP task data as it yields optimal results.

\paragraph{Instruction Reconstruction Task Ratio}
We explore the influence of the ratio of the IR task and thought generation task data on the model's performance using CodeLlama-7B as the base model.
We consider introducing 20\%, 40\%, 60\%, and 80\% proportions of IR task data and randomly mask 15\% of instruction clauses, prompting the model to reconstruct the masked parts.
As shown in subfigure (c) of Figure \ref{fig:ratio_data}, on evaluation benchmarks like MATH and AuQA, the model achieves optimal results when the task ratio $r_{task}=0.6$, while other datasets also achieve relatively good performance. 
Therefore, the IR task ratio is set to $r_{task}=0.6$.

\paragraph{Intermediate Reasoning State Masking Ratio}
We explore the influence of the masking ratio of intermediate reasoning states in the IRSP task on the performance of LLMs, based on CodeLlama-7B.
Building on the exploration of the optimal task ratio ($r_{task}=0.2$) for the IRSP task, 15\%, 40\%, 60\%, and 80\% proportions of intermediate reasoning steps are masked. 
The model is prompted to predict the masked reasoning states  by IRSP instructions.
As shown in subfigure (b) of Figure \ref{fig:ratio_data}, when the masking ratio $r_{mask}=0.15$, the model achieves optimal results on MATH, AQuA, and SVAMP, and performs well on other mathematical reasoning tasks. 
Hence, the intermediate reasoning state masking ratio is set to $r_{mask}=0.15$.

\paragraph{Instruction Clause Masking Ratio}
We explore the influence of the masking ratio of instruction clauses in the IR task on the model's reasoning capabilities and identifies superior hyper-parameter settings.
We consider masking 15\%, 40\%, 60\%, and 80\% proportions of instruction clauses in the instructions. 
The data ratio for the IR task and thought generation task is set to the previously determined optimal value, $r_{task}=0.6$.
As shown in Figure \ref{fig:ratio_data}, when the instruction clause masking ratio is set to $r_{mask}=0.6$, the model achieves optimal results on tasks like GSM8K and MATH, and performs relatively well on other tasks. 
Therefore, the IR task instruction clause masking ratio is set to $r_{mask}=0.6$.

\subsection{Case Study}
\begin{figure*}[t]
	\centering
	\includegraphics[width=6.2in]{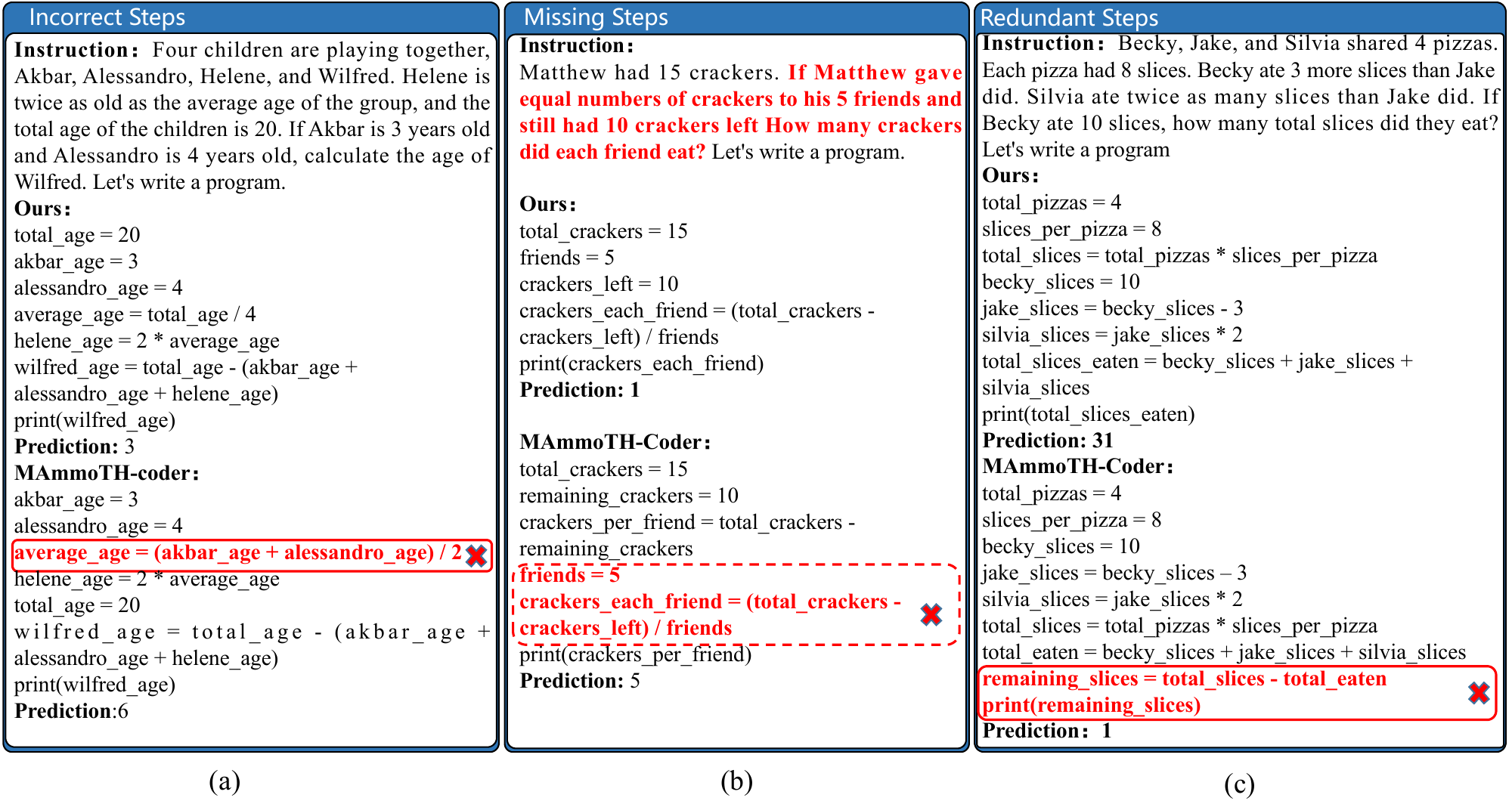}
	\caption{A case study on GSM8K and SVAMP datasets for comparison of baseline and our method}
 \label{fig:case_study}
\end{figure*}
This section compare the baseline model MAmmoTH-Coder and the proposed model, discussing the role of the IRSP and the IR task in addressing the quality issues of generated thoughts.

Figure \ref{fig:case_study} illustrates three instances extracted from GSM8K and SVAMP tasks, along with the predictions of the baseline and proposed models.
In the first instance, the baseline model erroneously calculates the average age of the group, leading to subsequent errors in Helene's age and, ultimately, Wilfred's age. 
In contrast, the proposed method correctly understands the semantics of \textit{"Helene is twice as old as the average age of the group"} and generates the correct steps to calculate the average age, resulting in the correct answer.
The second instance reveals that the baseline model loses the interpretation of the instruction clause \textit{"If Matthew gave equal numbers of crackers to his 5 friends and still had 10 crackers left"}. 
This leads to a missing inference step in the dashed box and subsequently results in an incorrect answer.
The third instance highlights that the baseline model outputs redundant inference steps in the last two steps that do not correspond to any instruction clause in the given instructions, resulting in an incorrect final answer.

These instances demonstrate that LLMs exhibit instruction comprehension errors and generate intermediate step errors, leading to error final answer. 
Such issues manifest as reasoning step errors, omissions, and redundancies during the thought generation process.
The proposed method aims to finely learn the mapping from the instruction space to the thought space. 
The IRSP task learns to parse each instruction clause to a accurate thought to enhance overall performance. 
The IR task, on the other hand, improves the model's understanding of instructions and the mapping between instructions and thoughts in the reverse direction.
In comparison with the aforementioned instances, the proposed method effectively alleviates issues such as reasoning step errors, omissions, and redundancies in the thought generation process of LLMs.

\subsection{Error Analysis}
To explore the bottlenecks in current LLMs when solving mathematical reasoning problems and discuss future research directions, this section conducts a statistical analysis of error instances in the evaluation tasks GSM8K and SVAMP.

\begin{figure*}[h]
	\centering
	\includegraphics[width=5.4in]{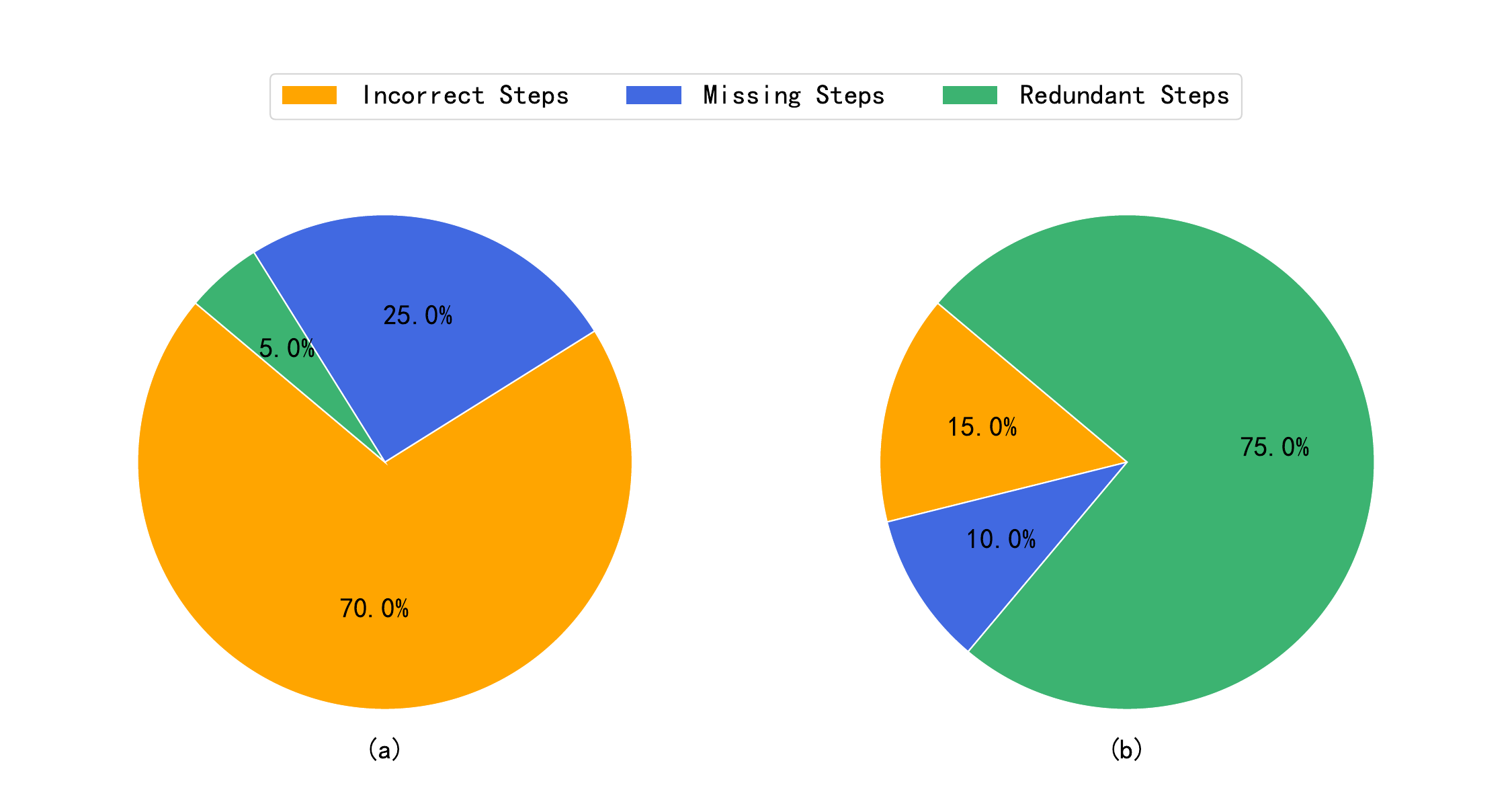}
	\caption{Errors instances from GSM8K(left) and SVAMP(right)}
 \label{error_statics}
\end{figure*}

\begin{figure*}[h]
	\centering
	\includegraphics[width=6in]{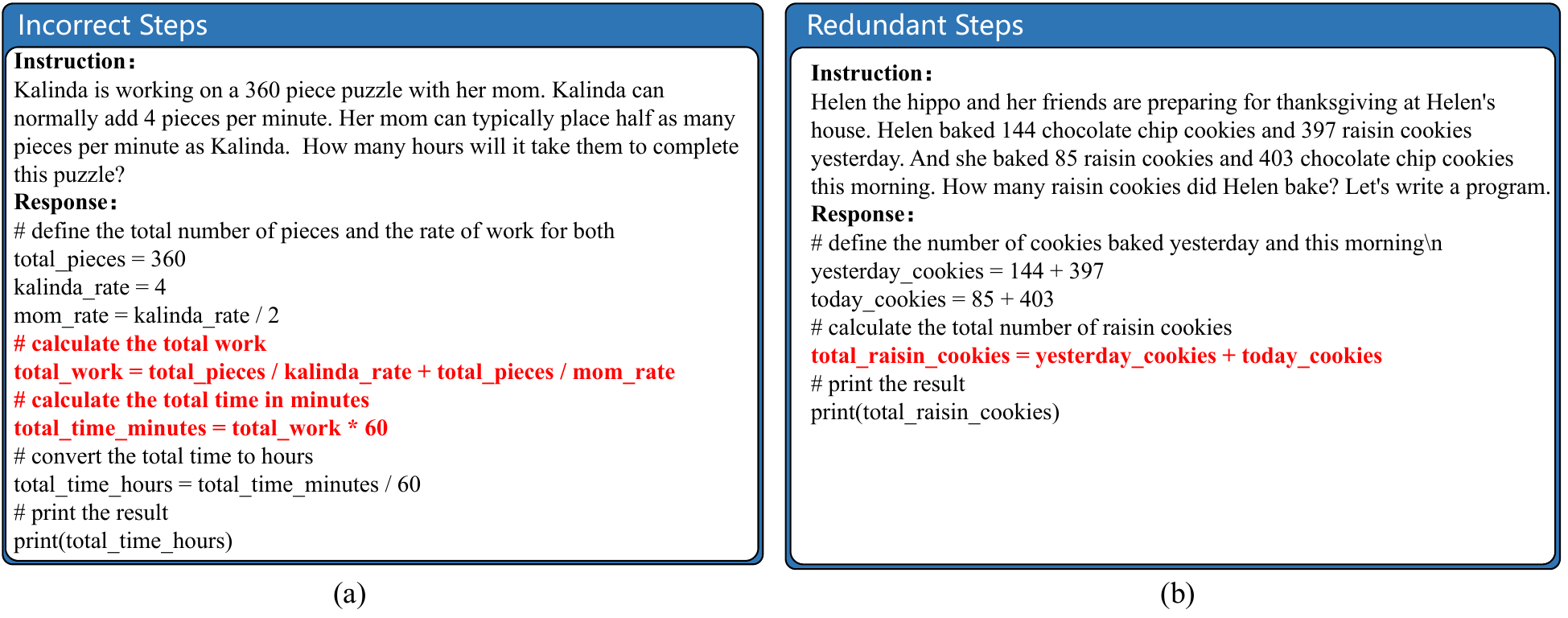}
	\caption{Analysis of instances related to errors and redundancies in reasoning steps}
 \label{error_examples}
\end{figure*}

As shown in Figure \ref{error_statics}, 20 error cases were randomly selected from the test sets of GSM8K and SVAMP, categorized based on error types such as missing, incorrect, and redundant reasoning steps.
In the in-domain task GSM8K, the predominant error type is reasoning step errors. 
Conversely, in the out-of-domain task SVAMP, reasoning step redundancy is the primary error type.
Through case analysis, it was observed that reasoning step errors mainly stem from misinterpreting some ambiguous instruction clauses, leading to the use of incorrect values or operators in generating the thoughts. 
For instance, as depicted in subfigure (a) of Figure~\ref{error_examples}, the bold red steps represent the erroneously generated thought, attributed to the model's failure to comprehend the shared puzzle-solving activity between Kalinda and Kalinda's mom, resulting in an incorrect prediction.

Reasoning step redundancy occurs due to the presence of instruction clauses unrelated to the calculation in the instructions, with the model habitually parsing and incorporating every instruction clause into the final prediction. 
As shown in subfigure (b) of Figure \ref{error_examples}, the problem only requires calculating the quantity of \textit{"raisin cookies"}, but the instructions contain clauses related to \textit{"chocolate chip cookies"}, causing the model to be easily disturbed by irrelevant information, leading to the generation of redundant thoughts that impact the final answer prediction.
While our approach alleviates some quality issues in thought generation, LLMs still face challenges in solving complex mathematical reasoning tasks, as evident in the discussed instances. Therefore, in future work, distinguishing ambiguous instructions and effectively identifying irrelevant instructions for improving numerical reasoning abilities is crucial.

\subsection{Related Work}
\paragraph{Chain-of-Thought and Variants}
Recent research explores various structures and forms of thoughts to enhance the model's planning and reasoning ability in handling complex problems.
For instance, Tree-of-Thought(ToT)~\cite{DBLP:journals/corr/abs-2305-10601} incorporates backtracking and search mechanisms in generating CoT.
Similarly, the method of cumulative reasoning~\cite{DBLP:journals/corr/abs-2308-04371} records all correct historical reasoning steps for exploration in the current search branch.
Graph-of-Thought~\cite{DBLP:journals/corr/abs-2308-09687} incorporates operations like correction and aggregation on thoughts.
Additionally, to mitigating the inherent computational drawbacks of LLMs, program-based thought generation approaches are proposed, such as Program-of-Thought(PoT)~\cite{DBLP:journals/corr/abs-2211-12588} and PAL~\cite{DBLP:journals/corr/abs-2211-10435}.

\paragraph{Instruction Tuning}
Instruction tuning offers significant advantages in enhancing model controllability, commonly utilized to guide the generation of data that aligns with expectations and constraints~\cite{DBLP:journals/corr/abs-2212-10560, DBLP:journals/corr/abs-2309-05653, DBLP:journals/corr/abs-2310-03731, DBLP:journals/corr/abs-2306-08568}.
Further, it is beneficial to elicit the capabilities of instruction understanding and execution via instruction tuning~\cite{DBLP:journals/corr/abs-2309-05653,DBLP:journals/corr/abs-2306-08568}.
Furthermore, harnessing the capacity to understand and execute instructions enables the execution of desired operations \cite{DBLP:journals/corr/abs-2205-10625,DBLP:conf/emnlp/Wang0S22,DBLP:conf/eacl/Chen23}. For example, the Least-to-Most method~\cite{DBLP:journals/corr/abs-2205-10625} is advocated for breaking down intricate problems into sub-problems and solving them progressively to obtain the answer.

\paragraph{LLMs for Mathematical Reasoning}
The prevailing approach for mathematical reasoning involves integrating LLMs with CoT and are categorized into:
(1) Enhancing the representation of reasoning steps: Past studies\cite{DBLP:journals/corr/abs-2201-11903,DBLP:journals/corr/abs-2305-10601,DBLP:journals/corr/abs-2308-09687,DBLP:journals/corr/abs-2211-12588,DBLP:journals/corr/abs-2211-10435} explored various forms of thoughts, such as chains, trees, graphs, and executable programs.
(2) Improving the generated reasoning steps, such as voting with multiple reasoning path~\cite{DBLP:journals/corr/abs-2203-11171}, decomposing complex problems~\cite{DBLP:journals/corr/abs-2203-11171,DBLP:journals/corr/abs-2205-10625,DBLP:conf/emnlp/Wang0S22} and correction-based methods~\cite{DBLP:journals/corr/abs-2110-14168,DBLP:journals/corr/abs-2212-09561}.  
(3) Exploring methods for obtaining extensive data: current approaches involve collecting high-quality public datasets or generating substantial instruction data~\cite{DBLP:journals/corr/abs-2308-09583} as well as data augmentation techniques~\cite{DBLP:journals/corr/abs-2309-12284}.
\end{document}